# A multimodal gesture recognition dataset for desktop human-computer interaction

Qi Wang*
Xidian University
Xi'an, China
qiwang0720@stu.xidian.edu.cn

Fengchao Zhu*
Xidian University
Xi'an, China
fczhu@stu.xidian.edu.cn

Guangming Zhu
Xidian University
Xi'an, China
gmzhu@stu.xidian.edu.cn

Liang Zhang
Xidian University
Xi'an, China
liangzhang@stu.xidian.edu.cn

Ning Li
China Institute of Marine Technology
& Economy
Beijing, China
lining199008@163.com

Eryang Gao
China Institute of Marine Technology
& Economy
Beijing, China
thugaoeryang@163.com

*Abstract*—Gesture recognition is an indispensable component of natural and efficient human-computer interaction technology, particularly in desktop-level applications, where it can significantly enhance people's productivity. However, the current gesture recognition community lacks a suitable desktop-level (top-view perspective) dataset for lightweight gesture capture devices. In this study, we have established a dataset named GR4DHCI. What distinguishes this dataset is its inherent naturalness, intuitive characteristics, and diversity. Its primary purpose is to serve as a valuable resource for the development of desktop-level portable applications. GR4DHCI comprises over 7,000 gesture samples and a total of 382,447 frames for both Stereo IR and skeletal modalities. We also address the variances in hand positioning during desktop interactions by incorporating 27 different hand positions into the dataset. Building upon the GR4DHCI dataset, we conducted a series of experimental studies, the results of which demonstrate that the fine-grained classification blocks proposed in this paper can enhance the model's recognition accuracy. Our dataset and experimental findings presented in this paper are anticipated to propel advancements in desktop-level gesture recognition research.

*Keywords—gesture recognition, dataset, top-view, desktop-level*

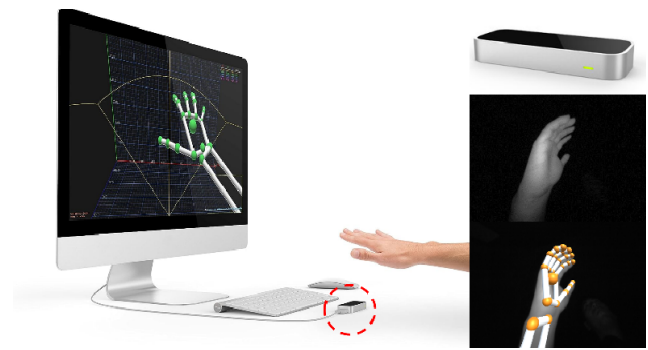

Fig. 1. (Left) A subject using Leap Motion and performs a gesture. (Right-top) The Leap Motion Controller we use. (Right-mid) The image captured by the Leap Motion Controller. (Right-bottom) skeleton detected by the Leap Motion Controller.

## I. INTRODUCTION

In the rapid development history of human-computer interaction technology, gesture-based interaction has consistently held a prominent position. Compared to traditional methods of interaction involving keyboards and mice, gesture interaction offers advantages such as contactless operation, natural intuitiveness, high efficiency, and low learning curve. Consequently, gesture recognition finds versatile applications in various domains and scenarios. For instance, it aids in facilitating daily communication between deaf and hearing individuals through gesture recognition. During times of pandemics, gesture recognition technology plays a pivotal role in reducing physical contact in public facilities. In the realm of motion-based gaming, gesture recognition enhances the immersive gaming experience.

Simultaneously, with the advancement of deep neural networks, more efficient gesture datasets [1-5] and gesture recognition algorithms have been proposed. These datasets encompass a variety of applications, including gesture data systems (NvGesture [1]) through multiple devices, first-person perspective gesture data collected via wearable collected for interaction with automotive continuous devices (EgoGesture [2]), large-scale datasets featuring continuous gestures representing symbols captured from a third-person perspective (ChaLearn ConGD [3]), datasets recorded from a distance for conference scenarios (LD-ConGR [4]), and datasets specifically tailored for computer interaction (Jester [5] and IPN Hand [6]). Within the desktop-level datasets, it has been observed that the majority of data is captured from a front-facing camera, necessitating users to raise their hands to higher positions. However, in everyday computer interaction scenarios, users often require the ability to perform gestures for extended periods. Sustaining a series of actions in a natural state for prolonged periods can be challenging, and the overlap between facial and hand regions can introduce difficulties in gesture recognition. These factors have hindered the widespread adoption of gesture-based interactions.

In this article, for the purpose of gesture recognition on the desktop using the Leap Motion Controller (LMC) [7], we have established a large-scale video dataset known as GR4DHCI[1]. This dataset comprises over 7,339 infrared video samples, each paired with corresponding skeletal sequences, resulting in a total frame count of 382,447 frames. The dataset's infrared and skeletal streams were captured at a rate of 30 frames per second, with the infrared images having dimensions of $640 \times 480$ pixels and containing 25 skeletal key points. Data was collected from volunteers of varying genders, heights, and body types, totaling 16 individuals. To meet the demands of prolonged usage without causing fatigue and to ensure naturalness of gestures, we placed particular emphasis on the naturalness of hand movements. As a result, we designed seven categories of dynamic gestures. Furthermore, due to the unique perspective provided by the LMC, our dataset encompasses gestures performed at differ-ent heights, various front-to-back depths, and different angles, substantially increasing the diversity of the data.

* Equal Contribution
This work was supported by the Chinese Defense Advance Research Program (50912020105).

[1] https://pan.baidu.com/s/1wo-7Bius9axORWEnxOvWvw?pwd=sx8u

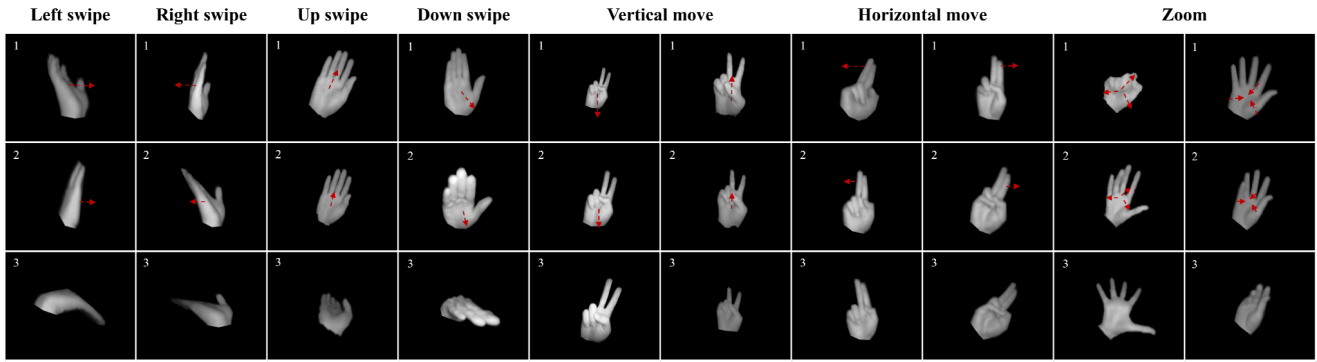

Fig. 2. Seven gesture classes of GR4DHCI dataset, including 7 gestures ('left swipe', 'right swipe', 'up swipe', 'down swipe', 'vertical move', 'horizontal move', 'zoom') gestures. Each column shows the general practice of the above gestures. The red arrow in the figure indicates the direction of hand movement.

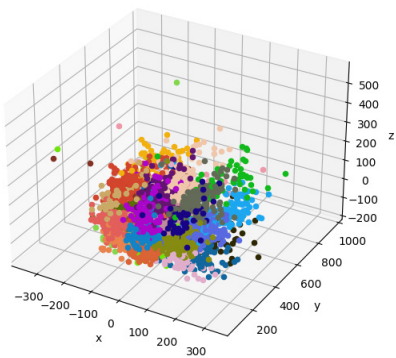

Fig. 3. 3D point cloud of our dataset drawn from palm positions. Different colors represent different positions, a total of 27 positions.

Subsequent to this, we conducted a series of experimental analyses using the proposed GR4DHCI dataset. Given the dual modality nature of our dataset, our experiments were conducted independently for the Stereo IR and skeletal modalities. Initially, we employed the topology-learnable graph convolution (TL-GCN) approach, as presented in [8], for skeleton-based gesture recognition. Additionally, for stereo-based gesture recognition, we adopted the convolutional LSTM (ConvLSTM) method proposed in [9]. As the results, we achieved accuracy rates of 93.50% and 82.57% for these respective modalities.

To harness the full potential of our dataset, we introduced a fine-grained classification block based on infrared imagery and skeletal displacement to enhance model recognition rates.

Within these blocks, we adopted a two-pronged approach. Firstly, we input each frame of the gesture video into a simplified MobileNet [10] image classification network to capture the finer details of hand shapes within the video. Secondly, we manually extracted geometric motion features from the skeletal data to represent distinct gesture's common geometric characteristics. Finally, we corrected the model using a fusion descriptor comprised of predefined hand shape information and geometric motion features. This strategic approach led to an increase in accuracy by 2.64% and 7.75% on [8] and [9].

## II. RELATED WORK

### A. Gesture Recognition Datasets

Within the publicly available datasets, there exists a diverse exists a diverse range of datasets that cover various perspectives. For instance, NvGesture [1] collected data from 20 volunteers in a simulated driving scenario using Soft Kinetic depth cameras and DUO 3D cameras, resulting in a dataset comprising 25 different hand gestures, including RGB, depth, and Stereo IR data. EgoGesture [2], on the other hand, gathered over 24,000 gesture samples from 50 distinct subjects using RealSense cameras mounted on headgear. ChaLearn ConGD [3] offers the largest repository of gesture data across different application domains, encompassing sign language, signals, and pantomime gestures, which were not explicitly designed for human-computer interaction purposes. LD-ConGR [4], on the other hand, summarizes the common characteristics of existing gesture datasets, emphasizing close proximity capture. It consolidates data from 30 participants across five different scenes, comprising 542 videos and 44,887 gesture instances, with color stream data boasting resolutions of up to $1280 \times 720$ pixels. Datasets like Jester [5] and IPN Hand [6] were explicitly gathered for computer interaction applications, with Jester featuring 27 gesture classes from 1,376 actors and IPN Hand containing over 4,000 gesture instances from 50 subjects. Notably, IPN Hand includes the highest number of consecutive gestures and the fastest intra-class variation among the datasets.

In practical applications, the choice of dataset often hinges on the perspective required. Based on the three perspectives defined in EgoGesture [2] for the gesture recognition domain, the first-person perspective involves a camera on the user's wearable device, commonly used in devices such as VR/AR glasses and headsets. Notable datasets for this perspective include EgoGesture [2], etc. The second-person perspective is characterized by a camera as the receiver, capturing intentional interactions from the performer facing the camera. Datasets such as NvGesture [1], Chalearn ConGD [3], Jester [5], IPN Hand [6], and LD-ConGR [4] mostly fall into this category. The third-person perspective treats the camera as an observer, capturing the performer's natural actions, akin to a surveillance perspective exemplified by MuViHand [32]. In accordance with these definitions, the gestures we use for LMC-based computer interaction fall under the second-person perspective. Among datasets in this perspective, NvGesture [1] offers a top view perspective similar to our work, while our dataset employs a down-view perspective, representing a vertically mirrored relationship. Both Jester and IPN Hand share usage scenarios and interaction styles with our dataset, all designed for desktop computer interaction in everyday scenarios. Consequently, our dataset can be categorized as a all designed for desktop computer interaction in everyday scenarios. Consequently, our dataset can be categorized as a second-person perspective, down-view dataset, specifically tailored for LMC desktop interaction. To the best of our knowledge, there is currently no dataset of this type available.

TABLE I.  STATISTICS OF THE PROPOSED GR4DHCI DATASET

| Gesture | Instance | | | Duration | | | |
|---|---|---|---|---|---|---|---|
| | Total | Train | Test | Avg. | Std. | Max. | Min. |
| Left swipe | 753 | 565 | 188 | 25.53 | 10.44 | 97 | 9 |
| Right swipe | 753 | 564 | 189 | 25.70 | 10.05 | 66 | 8 |
| Up swipe | 755 | 567 | 188 | 26.40 | 11.20 | 89 | 8 |
| Down swipe | 752 | 564 | 188 | 27.02 | 11.18 | 77 | 8 |
| Vertical move | 752 | 564 | 188 | 47.05 | 24.87 | 153 | 10 |
| Horizontal move | 752 | 564 | 188 | 46.47 | 22.97 | 118 | 9 |
| Zoom | 2822 | 2162 | 660 | 82.66 | 68.38 | 317 | 6 |
| Total | 7339 | 5550 | 1129 | 40.12 | 51.03 | 317 | 6 |

TABLE II.  COMPARISON OF OUR DATASET GR4DHCI AND POPULAR GESTURE RECOGNITION DATASETS

| Dataset | Classes | Video | Instances | Actors | Resolution | | | Frame rate/fps | View |
|---|---|---|---|---|---|---|---|---|---|
| | | | | | RGB | Stereo IR | Skeleton | | |
| Jester [5] | 27 | 148092 | 148092 | 1376 | √ | | | 12 | 2nd |
| NvGesture [1] | 25 | 1532 | 1532 | 20 | √ | √ | | 30 | 2nd |
| Chalearn ConGD [3] | 249 | 22,535 | 47,933 | 21 | √ | | | 10 | 2nd |
| IPN Hand [6] | 13 | 200 | 4218 | 50 | √ | | | 30 | 2nd |
| EgoGesture [2] | 83 | 2081 | 24161 | 50 | √ | | | 30 | 1st |
| Ours | 7 | 7339 | 7339 | 16 | | √ | √ | 30 | 2nd |

B. *Gesture Recognition Methods*

*1) Spatiotemporal networks:* Action recognition architectures are crafted through the expansion of image classification networks to incorporate a temporal dimension while retaining spatial characteristics. 3D ConvNets, as demonstrated in prior works [11-13], broaden the scope of 2D image models [14-17] into the spatiotemporal domain, maintaining consistent treatment of both spatial and temporal dimensions. LSTM/RNN models also exhibit significant advantages in handling long sequential data. ConvLSTM [18] takes 2D feature maps as input and explicitly encodes spatial correlations within both the input-to-state and state-to-state transitions. Therefore, ConvLSTM has been widely applied in action recognition [19][20], gesture recognition [21-23,9]. Action recognition also employs networks based on two-stream or 3D CNN architectures, including renowned models such as C3D [24], Res3D [25], I3D [13], SlowFast [26], X3D [27].

*2) Graph convolutional networks:* As human actions can be viewed as spatiotemporal graphs, in recent years, researchers have extended convolution techniques into the domain of graphs for skeleton-based action recognition [28-30]. 2s-AGCN [29] proposed a two-stream adaptive GCN over the body joints and bones. ST-GCN [31] proposed a novel model of dynamic skeletons. Inspired by the work of both 2s-AGCN and ST-GCN, TL-GCN proposed a topology-learnable graph convolution to fully use the self-learning ability of deep learning.

III. THE FREE-HAND DATASET

A. *Data Collection*

In order to collect the dataset, we choose Leap Motion Controller (LMC) as our desktop-level data collector due to its hardware size and integrating both Stereo IR and 3D skeleton modules. In these two modality datasets, Stereo IR stream is recorded in the resolution of 640×480 with the frame rate of 30 fps, the 3D skeleton data contains the 3D coordinates of 25 body joints per frame with the same frame rate as Stereo IR stream. As shown in Fig. 1, the subject is prompted by the acquisition software to make gestures freely and naturally. During the collection, the subjects are asked to execute gestures in 27 different poses which are defined as follows: Along table edge we set left-center-right; along the vertical desktop orientation we set low-medium-high; along the vertical table edge we set front-center-back. The distribution of gesture position shows in Fig. 3. Each gesture acquisition location is a combination of three directions. We endeavor to comprehensively simulate all possible hand gesture poses within our dataset to make it applicable to various usage scenarios.

Before collecting data, we first teach the subjects how to perform each gesture and tell them the gesture names (short descriptions). Then, we generate a gesture name list and a list of location with random order for each subject by the home-made data acquisition software. Therefore, the gesture name and gesture location are informed and the gesture can be performed accordingly. The standard practice for these gestures is shown in Fig. 2. According to the information which is generated, they were asked to make one gesture as a single session which was recorded as a video.

B. *Dataset Statistics*

The GR4DHCI dataset contains 7339 gesture instances of 7 different hand gesture classes. We randomly divided the dataset into training set and testing set by the subjects. The training dataset comprises 5550 gestures gathered from 12 subjects. The remaining 4 subjects contributed 1129 gestures,

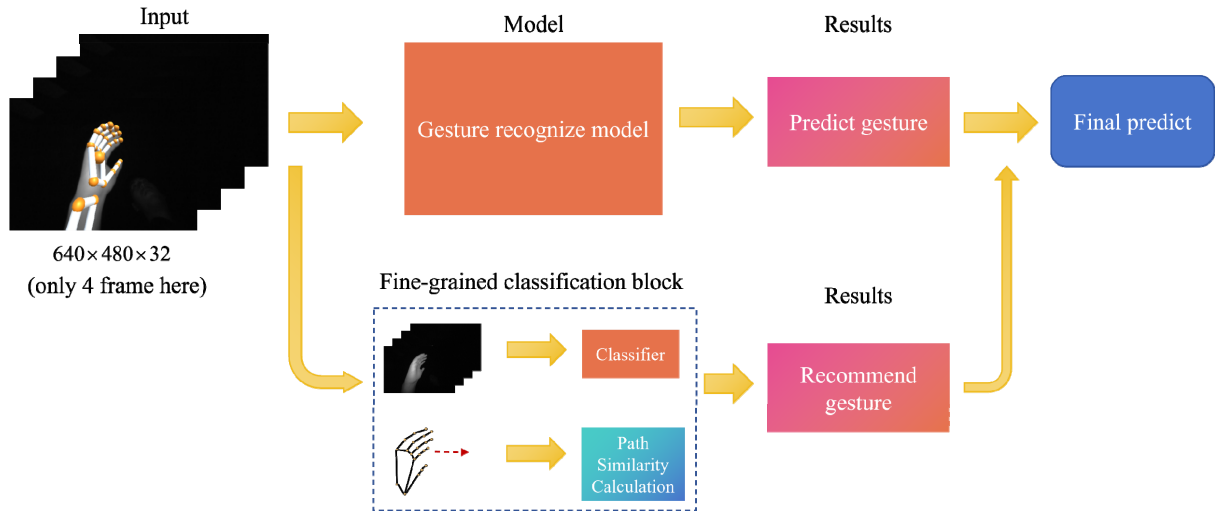

Fig. 4. Pipeline of our method. The input data is a multimodal data sequence. One of the multimodal data gets predicted gestures through its gesture recognition model. In addition, the input data passed through FGCB and obtain recommend gesture. Get the final gesture based on the recommended and predicted gestures.

TABLE III. RECOGNITION ACCURACY FOR THE DATASET GR4DHCI WITH FGCB

| Model | Input | Modality | Results |
|---|---|---|---|
| Res3D + ConvLSTM + MobileNet [9] | 32-frames | Stereo IR | 82.57% |
| TL-GCN [8] | 32-frames | Skeleton | 93.50% |
| Res3D + ConvLSTM + MobileNet + FGCB | 32-frames | Stereo IR + Skeleton | 90.32% |
| TL-GCN + FGCB | 32-frames | Stereo IR + Skeleton | 96.14% |

TABLE IV. RESULT OF REPRESENTATIVE METHOD ON THE STEREO IR MODALITY OF GR4DHCI

| Model | Input | Modality | Results |
|---|---|---|---|
| I3D [13] | 32-frames | Stereo IR | 87.76% |
| SlowFast [26] | 32-frames | Stereo IR | 89.97% |
| X3D [27] | 32-frames | Stereo IR | 89.70% |
| Res3D + ConvLSTM + MobileNet [9] | 32-frames | Stereo IR | 82.57% |
| Res3D + ConvLSTM + MobileNet + FGCB (ours) | 32-frames | Stereo IR | 90.32% |

forming the testing dataset. The frequency of each gesture class is presented in Table I.

There are more "zoom" gestures here as we have combined "zoom in" and "zoom out" gestures for more natural zoom applications. In regard to the "move" gesture, as illustrated in Table I, we have established two distinct forms: "vertical movement" and "horizontal movement". With the acquisition of these two movement gestures, we have encompassed two directional aspects. To elaborate, "vertical movement" encompasses both the upward and downward directions and is captured continuously. This offers the distinct advantage of providing finer-grained insights throughout the transition process, particularly beneficial for applications involving mobile gesture control. It is evident that the duration of gestures exhibits substantial variability, both within the same gesture class and across different gesture classes. In addition to the special gestures of moving and zooming, the discrepancy in duration between the longest and shortest gestures can be as high as 89 frames (97 frames compared to 8 frames). Even in instances belonging to the same gesture class, the maximum disparity in duration is 82 frames ("left swipe" class, 97 frames compared to 9 frames). The significant variation in gesture duration can be attributed to two primary factors. Firstly, individuals execute gestures at distinct speeds, and secondly, diverse gesture categories exhibit varying temporal requirements. This substantial dissimilarity and the inherent uncertainty in gesture duration pose formidable challenges to the field of gesture recognition.

In Table II, we compare our dataset GR4DHCI with the gesture recognition datasets that are currently publicly available. In all these datasets, our dataset GR4DHCI is established for down view desktop-level gesture recognition. In the previously published datasets, the subject performs gestures in a less natural and comfortable position, significantly impacting both the bodily sensation and the efficiency of gesture execution. Considering our dataset, subject can use gestures freely in a very natural state. Besides, our dataset comprises both Stereo IR and Skeleton stream two modalities as the Jester [5], Chalearn ConGD [3], IPN hand [6] and EgoGesture [2] provides RGB only. Moreover, our data are captured at a high frame rate (30 fps).

IV. EXPERIMENTAL STUDIES

A. Fine-grained classification block

During the dataset creation process, we observed that more natural gestures often exhibit some common patterns. For instance, a leftward waving gesture can be decomposed into a displacement of a waving hand along a certain path. Leveraging this characteristic, we introduced a fine-grained classification block (FGCB).

As FGCB comprises two main components: infrared image classification and skeleton similarity comparison as shown in Fig. 4. In the infrared image classification component, we selected MobileNet [10] as our classifier and defined the classes as "leftward/rightward wave", "upward/downward wave", "move", and "zoom". The data for these classes were annotated by cropping from the GR4DHCI videos. Then, we follow the training procedure aligns with the steps outlined in [10]. For the skeleton path extraction, we predefined the paths for each class. During the recognition process, we first obtained the gesture recognition result through the gesture recognition network. Subsequently, we input the result and two modalities' data into FGCB, and classify the image sequence frame by frame. Secondly, we assume that $C_1, C_2, ..., C_n$ is the hand type recognition result of the image sequence after the classification model, and let $N$ be the sequence length which we use in the experiment is 32. We can get the category with the highest probability of occurrence as the sequence prediction result:

$$Class = \arg\max_{i=1}^{n} Count(C_i) \quad (1)$$

On the other hand, we gave the current stream's skeleton coordinate sequences as $S_{now}$ and the predefined trajectory sequence as $S_{pre}$, with lengths $N$ and $M$, we compute the similarity between the two sequences using the Edit Distance on Real sequence (EDR) [33] described as:

$$LCSS(A,B) = \begin{cases} 0, \text{if } n=0 \text{ or } m=0 \\ 1 + LCSS(Rest(S_{now}), Rest(S_{pre})), \text{if } d(Head(S_{now}), Head(S_{pre})) \leq \varepsilon \\ \max \begin{cases} LCSS(S_{now}, Rest(S_{pre})) \\ LCSS(Rest(S_{now}), S_{pre}) \end{cases}, \text{otherwise} \end{cases} \quad (2)$$

$$Similar(\delta, \varepsilon, A, B) = \frac{LCSS(A,B)}{\min(n,m)} \quad (3)$$

Here, $\varepsilon$ represents the matching distance threshold, and $\delta$ represents the interval threshold.

Lastly, if the similarity exceeds a predefined threshold of 0.7, then we can get the final predict. If the recommended gesture differs from the gesture recognition model's result, we choose the recommended gesture as the correct gesture. Although this approach is aimed at enhancing recognition accuracy, it's crucial to note that a prerequisite for this module is that FGCB must exhibit higher accuracy than the gesture recognition network to effectively elevate the overall recognition accuracy.

### B. Experimental setup

In the training phase, we first cut or fill the frame length of the gesture to a fixed length of 32 frames which can augment the data at the same time, scale the size of the Stereo IR data to 112×112 and reshape bone data into 3D vertex sequences. In the test phase of the gesture recognition model, we use the sliding window method to shape the test data to the input size of the model (in the experiment, the stride of the sliding window is set to 3), if the results of two consecutive predictions are the same, then the result is recognized as the final recognition result.

By using the FGCB, we obtained the results shown in Table III. Among them, the Res3D + ConvLSTM + MobileNet [9] trained by Stereo IR stream achieved an accuracy rate of 82.57%, and the TL-GCN [8] trained in the skeleton stream achieved an accuracy rate of 93.50%. As we can see, the training results of our dataset by skeleton are higher than the Stereo IR. We analyze the reason and conclude that the image sequence contains more background, which has more noise than the skeleton data. Moreover, it should be noted that after combining the unimodally trained model with FGCB, their accuracy increased by 7.75% (82.57% vs. 90.32%) and 2.64% (93.50% vs. 96.14%), such results sug-
gest that the hand features included in the Stereo IR stream and the route features of the skeletal modality can provide additional details to assist in recognition.

### C. State-of-the-art Evaluation

Table IV shows the results of state-of-the-art gesture and action recognition models on our proposed GR4DHCI dataset. Since repretraining the model may lead to performance differences, we used a unified pretraining model in the experiment, and in order to avoid the influence caused by the modality difference, we only used the data of the Stereo IR modality for comparison. As shown in the Table IV, we chose Res3D + ConvLSTM + MobileNet [9] using the Fine-grained classification block to compare with I3D [13], SlowFast [26] and X3D [27], and the experiments show that our method can be used in the proposed GR4DHCI performed well and provided a good baseline for this dataset.

## V. CONCLUSION

In this paper, we propose GR4DHCI, a large-scale bi-modal video dataset. This is the first dataset for desktop-level (top-down view) gesture recognition. GR4DHCI includes gestures based on natural and efficient movements, and considers the needs of people for different positions and angles when applying on the desktop. It includes 27 different positions, and the duration of gestures also varies greatly, which is sufficient diversity and authenticity. To take full advantage of the advantages of both modalities in our dataset, we propose a fine-grained classification block to classify and match skeletal trajectories for each image frame, further improving the original gesture recognition algorithm. In addition, we also conduct experiments and comparisons on GR4DHCI with representative methods for gesture and action recognition. It is hoped that our dataset can contribute in the field of desktop-level human-computer interaction and gesture recognition applications.